\documentclass[runningheads]{llncs}
\usepackage{graphicx}

\usepackage{tikz}
\usepackage{comment}
\usepackage{amsmath,amssymb} 
\usepackage{color}

\usepackage[accsupp]{axessibility}  

\usepackage{amsmath,amsfonts,bm}

\def\eqref#1{equation~\ref{#1}}

\def\1{\bm{1}}

\newcommand{\test}{\mathcal{D_{\mathrm{test}}}}

\def\vx{{\bm{x}}}

\def\mA{{\bm{A}}}

\DeclareMathAlphabet{\mathsfit}{\encodingdefault}{\sfdefault}{m}{sl}
\SetMathAlphabet{\mathsfit}{bold}{\encodingdefault}{\sfdefault}{bx}{n}

\def\gO{{\mathcal{O}}}

\def\sP{{\mathbb{P}}}

\def\sS{{\mathbb{S}}}

\def\emA{{A}}

\newcommand{\R}{\mathbb{R}}

\usepackage[capitalize]{cleveref}
\crefname{section}{Sec.}{Secs.}
\Crefname{section}{Section}{Sections}
\Crefname{table}{Table}{Tables}
\crefname{table}{Tab.}{Tabs.}

\begin{document}
\pagestyle{headings}
\mainmatter
\def\ECCVSubNumber{7522}  

\title{Shap-CAM: Visual Explanations for \\ Convolutional Neural 
Networks based on Shapley Value} 

\titlerunning{Shap-CAM}
%
\author{Quan Zheng\inst{1,2,3} \and Ziwei Wang\inst{1,2,3} \and
Jie Zhou\inst{1,2,3} \and Jiwen Lu\inst{1,2,3}}
\authorrunning{Zheng et al.}
%
\institute{Department of Automation, Tsinghua University, China \and
State Key Lab of Intelligent Technologies and Systems, China \and
Beijing National Research Center for Information Science and Technology, China \\
\email{\{zhengq20,wang-zw18\}@mails.tsinghua.edu.cn}
\email{\{jzhou,lujiwen\}@tsinghua.edu.cn}}
\maketitle

\begin{abstract}
Explaining deep convolutional neural networks has been recently 
drawing increasing attention since it helps to understand the networks' 
internal operations and why they make certain decisions.
Saliency maps, which emphasize salient regions largely connected to the 
network's decision-making, are one of the most common ways for 
visualizing and analyzing deep networks in the computer vision community. 
However, saliency maps generated by existing methods cannot represent authentic
information in images due to the unproven proposals about the weights of 
activation maps which lack solid theoretical foundation and fail to consider
the relations between each pixels.
In this paper, we develop a novel post-hoc visual 
explanation method called Shap-CAM based on class activation mapping. 
Unlike previous gradient-based approaches, Shap-CAM gets 
rid of the dependence on gradients by obtaining the importance of each 
pixels through Shapley value. We demonstrate that Shap-CAM achieves better
visual performance and fairness for interpreting the decision making 
process. Our approach outperforms previous methods on both recognition 
and localization tasks.
\keywords{CNNs, Explainable AI, 
Interpretable ML, Neural Network Interpretability}
\end{abstract}

\begin{figure}[t]
\centering
\includegraphics[width=\linewidth]{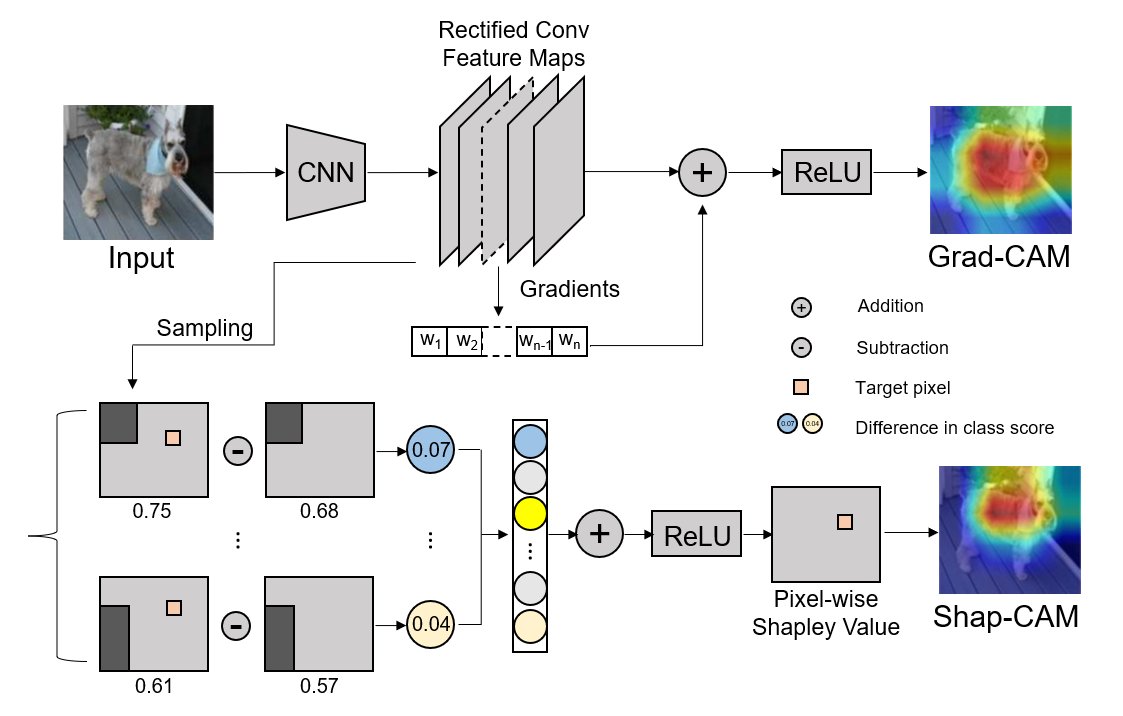}
\caption{Comparison between the conventional gradient-based CAM methods and
our proposed Shap-CAM. 
The gradient-based CAM methods, taking Grad-CAM for example,
combine the rectified convolutional feature maps and the gradients via 
backpropagation to compute the saliency map which 
represents where the model has to look to make the particular decision.
Our Shap-CAM introduces Shapley value to estimate
the marginal contribution of pixels. For a given pixel in the feature map
(viewed in color), 
we sample various pixel combinations and compute the score difference if the
given pixel is added. The score differences are synthesized to obtain
pixel-wise Shapley value, which generates the finally Shap-CAM saliency map.}
\label{fig:pipeline}
\end{figure}

\section{Introduction}
\label{sec:intro}

The dramatic advance of machine learning within the form of deep 
neural networks has opened up modern Artificial Intelligence (AI) 
capabilities in real-world applications.
Deep learning models achieve impressive results in tasks like object 
detection, speech recognition, machine translation, which offer tremendous
benefits.
However, the connectionist approach of deep learning is
fundamentally different from earlier AI systems where the
predominant reasoning methods are logical and symbolic.
These early systems can generate a trace of their inference
steps, which at that point serves as the basis for explanation. On the
other hand, the usability of today's intelligent systems is
limited by the failure to explain their decisions to human
users. This issue is particularly critical for risk-sensitive
applications such as security, clinical decision support or
autonomous navigation.

For this gap, various methods have been proposed by
researchers over the last few years to figure out what knowledge is hidden
in the layers and connections when utilizing deep learning models. 
While encouraging development has been carrying this field forward, 
existing efforts are restricted and the goal of explainable deep learning 
still has a long way to go, 
given the difficulty and wide range of issue scopes. 

In the context of understanding Convolutional Neural
Networks (CNNs), 
Zhou et al. proposed a technique called CAM (Class Activation Mapping), and 
demonstrated that various levels of the CNN functioned as unsupervised 
object detectors \cite{zhou2016learning}.
They were able to obtain heat maps that illustrate which regions of 
an input image were looked at by the CNN for assigning a label 
by employing a global average pooling layer and showing the weighted 
combination of the resulting feature maps at the penultimate 
(pre-softmax) layer. However,
this technique was architecture-sensitive and involved retraining 
a linear classifier for
each class. Similar methods were examined with different
pooling layers such as global max pooling and log-sum-exp pooling
\cite{oquab2015is,pinheiro2015from}. 
After that, Selvaraju et al. developed Grad-CAM, an efficient version of 
CAM that combines the class-conditional property of CAM with current 
pixel-space gradient visualization techniques like Guided Back-propagation 
and Deconvolution to emphasize fine-grained elements on the image
\cite{selvaraju2016gradcam}. 
Grad-CAM improved the transparency of CNN-based models by displaying 
input regions with high resolution details that are critical for prediction. 
The variations of Grad-CAM, such as Grad-CAM++ \cite{chattopadhay2018gradcam++},
introduce more reliable expressions for pixel-wise weighting of the gradients.
However, gradient-based CAM methods cannot represent authentic
information in images due to the unproven proposals about the weights of 
activation maps
\cite{adebayo2018local,adebayo2018sanity,chang2018explaining,dabkowski2017real,lin2013network,omeiza2019smooth}.
Adversarial model manipulation methods fool the 
explanations by manipulating the gradients without noticeable modifications 
to the original images \cite{heo2019fooling}, proving that the 
gradient-based CAM methods are not robust and reliable enough.

To this end, Wang et al. proposed Score-CAM which got rid of the dependence 
on gradients by obtaining the weight of each activation map through
its forward passing score on target class \cite{wang2020scorecam}.
Though Score-CAM discarded gradients for generating explanations,
it still suffered from self designed expression of score
which lacked solid theoretical foundation and failed to take the 
relationship between pixels into consideration.
In this work, we present
a new post-hoc visual explanation method, named Shap-CAM, 
where the importance of 
pixels is derived from their marginal contribution to
the model output utilizing Shapley value.
Our contributions are:
\begin{itemize}
\item We propose a novel gradient-free visual explanation method, Shap-CAM, 
which introduces Shapley value in the cooperative game theory to estimate
the marginal contribution of pixels.
Due to the superiority of Shapley value and the consideration of 
relationship between pixels, more rational and accurate contribution of 
each pixel is obtained.
\item We quantitatively evaluate the generated saliency
maps of Shap-CAM on recognition and localization tasks and show that 
Shap-CAM better discovers important features.
\item We show that in a constrained teacher-student setting, 
it is possible to achieve an improvement in the performance
of the student by using a specific loss function
inspired from the explanation maps generated by
Shap-CAM, which indicates that our explanations discover authentic 
semantic information mined in images.
\end{itemize}

The remainder of the paper is organized as follows. In \cref{sec:related},
we introduce the related work about visual explanations and Shapley value. 
In \cref{sec:approach}, we
develop our Shap-CAM for the generation of visual explanations based on
Shapley value.
In \cref{sec:experiment}, we present some experimental results 
on recognition and localization tasks and show the effectiveness of our
proposed method. We finish the paper with final conclusions and remarks.

\section{Related Work}
\label{sec:related}

\subsection{Visual Explanations}
\label{sec:explanation}

We give a summary of related attempts in recent years 
to understand CNN predictions in this part. 
Zeiler et al. provided one of the earliest initiatives in this field, 
developing a deconvolution approach to better grasp what the higher 
layers of a given network have learned \cite{zeiler2014visualizing}. 
Springenberg et al. extended this work to guided backpropagation, 
which allowed them to better comprehend the impact of each neuron in a 
deep network on the input image \cite{springenberg2014striving}. 
From a different perspective, Ribeiro et al. introduced
LIME (Local Interpretable Model-Agnostic Explanations), 
an approach that uses smaller interpretable classifiers like sparse 
linear models or shallow decision trees to make a local approximation 
to the complex decision surface of any deep model \cite{ribeiro2016why}.  
Shrikumar et al. presented DeepLift, which approximates the instantaneous 
gradients (of the output with respect to the inputs) with discrete 
gradients to determine the relevance of each input neuron for a given 
decision \cite{shrikumar2017learning}.
Al-Shedivat et al. presented Contextual Explanation Networks (CENs), 
a class of models that learns to anticipate and explain its decision 
simultaneously \cite{shedivat2017contextual}. 
Unlike other posthoc model-explanation tools, CENs combine deep networks 
with context-specific probabilistic models to create explanations in 
the form of locally-correct hypotheses.

Class Activation Mapping (CAM) \cite{zhou2016learning} 
is a technique for discovering discriminative regions and giving 
understandable explanations of deep models across domains. 
In CAM, the authors 
demonstrate that a CNN with a Global Average Pooling (GAP) layer after the 
last convolutional layer shows localization capabilities despite not being 
explicitly trained to do so. The CAM explanation regards the importance of 
each channel as the weight of fully connected layer connecting the global 
average pooling and the output probability distribution. However, an 
obvious limitation of CAM is the requirements of a GAP penultimate layer 
and retraining of an additional fully connected layer. 
To resolve this problem, Grad-CAM \cite{selvaraju2016gradcam} extends 
the CAM explanation and regards the importance of each channel as the 
gradient of class confidence w.r.t. the activation map. In Grad-CAM, 
the authors naturally regard gradients as the importance of each channel towards
the class probability, which avoids any retraining or model modification.
Variations of Grad-CAM, like Grad-CAM++ \cite{chattopadhay2018gradcam++}, 
use different combinations of gradients and revise the weights for
adapting the explanations to different conditions.

However, gradient-based CAM methods do not have solid theoretical foundation
and receive poor performances when the gradients are not reliable.
Hoe et al. explored whether the neural network interpretation methods can be 
fooled via adversarial model manipulation, a model fine-tuning step that aims to
radically alter the explanations without hurting the accuracy of the original 
models \cite{heo2019fooling}. 
They showed that the state-of-the-art gradient-based interpreters
can be easily fooled by manipulating the gradients with no noticeable 
modifications to the original images, proving that the 
gradient-based CAM methods are not robust and reliable enough. 
Score-CAM gets rid of the dependence of gradients 
and introduces channel-wise increase of confidence as the importance of each 
channel \cite{wang2020scorecam}. 
Score-CAM obtains the weight of each activation map through
its forward passing score on target class, the final result
is obtained by a linear combination of weights and activation maps.
This approach however suffers from self-designed expression of score
which lacked solid theoretical foundation.
Besides, it fails to consider the relationship between different pixels.

\subsection{Shapley Value}

One of the most important solution concepts in cooperative
games was defined by Shapley \cite{shapley1953value}. This solution 
concept is now
known as the Shapley value. The Shapley value is useful when there
exists a need to allocate the worth that a set of players can achieve
if they agree to cooperate.
Although the Shapley value has been widely studied from a theoretical 
point of view, the problem of its calculation still exists. In
fact, it can be proved that the problem of computing the Shapley
value is an NP-complete problem \cite{deng1994on}.

Several authors have been trying to find algorithms to calculate
the Shapley value precisely for particular classes of games. In Bilbao
et al. for example, where a special class of voting game is examined, 
theoretical antimatriod concepts are used to polynomially
compute the Shapley value \cite{fernandez2002generating}. 
In Granot et al. a polynomial algorithm is developed for a special case 
of an operation research game \cite{granot2002cost}.
In Castro et al., it is proved that the Shapley value for an airport
game can be computed in polynomial time by taking into account
that this value is obtained using the serial cost sharing 
rule \cite{castro2008polynomial}. 

Considering the wide application of game theory to
real world problems, where exact solutions are often not possible, a
need exists to develop algorithms that facilitate this approximation.
Although the multilinear extension defined by Owen is an exact method 
for simple games \cite{owen1972multilinear},
the calculation of the corresponding integral is not a 
trivial task. So, when this integral
is approximated (using the central limit theorem) this methodology
could be considered as an approximation method.
In Fatima et al. , a randomized polynomial method for determining 
the approximate Shapley value is presented for 
voting games \cite{fatima2006analysis}. 
Castro et al. develop an efficient algorithm that can estimate the
Shapley value for a large class of games \cite{castro2009polynomial}.
They use sampling to estimate the Shapley value and
any semivalues. These estimations are efficient if the worth of any
coalition can be calculated in polynomial time.

In this work, we propose
a new post-hoc visual explanation method, named Shap-CAM, where the importance of 
pixels is derived from their marginal contribution to
the model output utilizing Shapley value.
Due to the superiority of Shapley value and the consideration of 
relationship between pixels, more rational and accurate explanations 
are obtained.

\section{Approach}
\label{sec:approach}

In this section, we first present the prelimilaries of visual explanations
and the background the CAM methods.
Then we introduce the theory of Shapley Value \cite{shapley1953value}, 
a way to quantify the marginal contribution of each player in the 
cooperative game theory. We apply this theory to our problem and propose
the definition of Shap-CAM. 
Finally, we clarify the estimation of Shapley value in our method.
The comparison between the conventional gradient-based CAM methods and 
our proposed Shap-CAM is illustrated in \cref{fig:pipeline}.

\subsection{Prelimilaries}

Let function $ \bm Y = f(\bm X) $ be a CNN which takes $\bm X$ as an input 
data point
and outputs a probability distribution $\bm Y$. We denote $Y^c$ as the
probability of class $c$. For the last convolutional layer, $\mA^k$ denotes
the feature map of the $k$-th channel.

In CAM, the authors 
demonstrate that a CNN with a Global Average Pooling (GAP) layer after the 
last convolutional layer shows localization capabilities despite not being 
explicitly trained to do so. However, an obvious limitation
of CAM is the requirements of a GAP penultimate layer and retraining of an
additional fully connected layer.
To resolve this problem, Grad-CAM \cite{selvaraju2016gradcam} 
extends the CAM explanation and regards the importance of each channel 
as the gradient of class 
confidence $\bm Y$ w.r.t. the activation map $\mA$, which is defined as:
\begin{equation}
\label{eq:grad-cam}
L_{ij,\ Grad-CAM}^c = ReLU \Big( \sum_k {w_k^c \emA_{ij}^k} \Big)
\end{equation}
where
\begin{equation}
\label{eq:grad-weight}
w_k^c = \frac{1}{Z} \sum_i \sum_j {\frac{\partial Y^c}{\partial \emA_{ij}^k}}
\end{equation}
Constant $Z$ stands for the number of pixels in the activation map. In 
Grad-CAM, 
the explanation is a weighted summation of the activation maps $\mA^k$,
where the gradients are regarded as the importance of each channel towards
the class probability, which avoids any retraining or model modification.
Variations of Grad-CAM, like Grad-CAM++ \cite{chattopadhay2018gradcam++}, 
use different combinations of gradients and revise $w_k^c$ in 
\cref{eq:grad-cam} for adapting the explanations to different conditions.

However, gradient-based CAM methods do not have solid theoretical foundation
and can be easily fooled by adversarial model manipulation methods 
\cite{heo2019fooling}. Without noticeable modifications to the original 
images or hurting the accuracy of the original models, the gradient-based 
explanations can be radically alteredby manipulating the gradients, proving 
that the gradient-based CAM methods are not robust and reliable enough. 
Score-CAM \cite{wang2020scorecam} gets rid of the dependence of gradients 
and introduces channel-wise increase of confidence as the importance score
of each channel. 
This approach however suffers from self designed expression of the score
which fails to consider the relationship between different pixels.

\subsection{Definition of Shap-CAM}

In order to obtain more accurate and rational estimation of the marginal 
contribution of each pixel to the model output, we turn to the cooperative
game theory. The Shapley value \cite{shapley1953value} is useful when there
exists a need to allocate the worth that a set of players can achieve
if they agree to cooperate.
Consider a set of $n$ players $\sP$ and a function $f(\sS)$ which represents 
the worth of the subset of $s$ players $\sS \subseteq \sP$. The 
function $f: 2^{\sP} \rightarrow \R$ maps each subset to a real number, where
$2^{\sP}$ indicates the power set of $\sP$. 
Shapley Value is one way to quantify the marginal contribution of each 
player to the result $f(\sP)$ of the game when all players participate. 
For a given player $i$, its Shapley value can be computed as:
\begin{equation}
\label{eq:shapley}
\begin{split}
Sh_i(f) = \sum_{\sS \subseteq \sP, i \notin \sS} \frac{(n-s-1)!s!}{n!}
\big[ f(\sS \cup \{i\}) - f(\sS) \big] 
\end{split}
\end{equation}
The Shapley value for player $i$ defined above can be interpreted as the average 
marginal contribution of player $i$ to all possible coalitions $\sS$ that can be 
formed without it. 
Notably, it can be proved that Shapley value is the only way 
of assigning attributions to players that satisfies the following four
properties:

\textbf{Null player.}
If the class probability does not depend on any pixels, then its attribution
should always be zero. It ensures that a pixel has no contribution if it does not
bring any score changes to every possible coalition.

\textbf{Symmetry.}
If the class probability depends on two pixels but not on their order 
(i.e. the values of the two pixels could be swapped, never affecting the 
probability), then the two pixels receive the same attribution.
This property, also called anonymity , is arguably a desirable property for
any attribution method: if two players play the exact same
role in the game, they should receive the same attribution.

\textbf{Linearity.}
If the function f can be seen as a linear combination of the
functions of two sub-networks (i.e. $f = a f_1 + b f_2$),
then any attribution should also be a linear combination, with the same
weights, of the attributions computed on the sub-networks,
i.e. $Sh_i(\vx|f) = a \cdot Sh_i(\vx|f_1) + b \cdot Sh_i(\vx|f_2)$.
Intuitively, this is justified by the need for preserving linearities 
within the network.

\textbf{Efficiency.}
An attribution method satisfies
efficiency when attributions sum up to the difference 
between the value of the function evaluated at the input,
and the value of the function evaluated at the baseline,
i.e. $\sum_{i=1}^n Sh_i = \Delta f = f(\vx) - f(\bm 0)$. 
In our problem, this property 
indicates that all the attributions of the pixels sum up to the difference
between the output probability of the original feature map and the output of 
the feature map where no original pixels remain. 
This property, also called completeness or conservation, has
been recognized by previous works as desirable to ensure
the attribution method is comprehensive in its accounting.
If the difference $\Delta f > 0$, there must exist some pixels assigned a 
non-zero attribution, which is not necessarily true for gradient-based
methods.

Back to our problem on class activation mapping, we consider each pixel $(i,j)$
in the feature map of the last convolutional layer $\mA$ as a player in the
cooperative game. Let $\sP = \{ (i,j) | i=1,\dots,h; j=1,\dots,w \}$ be the set
of pixels in the feature map $\mA$, where $h, w$ stand for the height and width
of the feature map. Let $n = h \cdot w$ be the number of pixels in the activation 
map. We then define the worth function $f$ in \cref{eq:shapley} as the class
confidence $Y^c$, where $c$ is the class of interest. For each subset $\sS 
\subseteq \sP$, $Y^c(\sS)$ represents the output probability of class $c$ when 
only the pixels in the set $\sS$ remain and the others are set to the average
value of the whole feature map. By the symbolization above, the original
problem turns to an $n$-player game $(\sP, Y^c)$. Naturally, the Shapley Value
of the pixel $(i,j)$ represents its marginal contribution to the class confidence.
Thus, we define the Shapley Value as the saliency map of our Shap-CAM:
\begin{equation}
\label{eq:shap-cam}
\begin{split}
&L_{ij,\ Shap-CAM}^c = Sh_{(i,j)}(Y^c) \\
&= \sum_{\sS \subseteq \sP, (i,j) \notin \sS} \frac{(n-s-1)!s!}{n!}
\big[ Y^c(\sS \cup \{(i,j)\}) - Y^c(\sS) \big]
\end{split}
\end{equation}
The obtained heatmap is then upsampled to the size of the original image, as it 
is done in Grad-CAM. 

The contribution formula that uniquely satisfies all these properties is that
a pixel's contribution is its marginal contribution to the class confidence 
of every subset of the original feature map.
Most importantly, this formula takes into account the interactions between 
different pixels. As a simple example, suppose there are two pixels that improve 
the class confidence only if they are both present or absent and harm the 
confidence if only one is present. The equation considers all these possible 
settings. This is one of the few methods that take such 
interactions into account and is inspired by similar approaches in Game
Theory.
Shapley value is introduced as an equitable way of
sharing the group reward among the players where equitable means satisfying 
the aforementioned properties. It's possible to make a direct
mapping between our setting and a cooperative game;
therefore, proving the uniqueness of Shap-CAM.

\subsection{Estimation of Shapley Value}

Exactly computing \cref{eq:shapley} would require $\gO(2^n)$ evaluations.
Intuitively, this is required to evaluate
the contribution of each activation with respect to all possible subsets 
that can be enumerated with the other ones. Clearly, the exact computation 
of Shapley values is computationally unfeasible for real problems.
Sampling is a process
or method of drawing a representative group of individuals or cases
from a particular population. Sampling and statistical inference are
used in circumstances in which it is impractical to obtain information 
from every member of the population. Taking this into account,
we use sampling in this paper to estimate the Shapley value and
any semivalues. These estimations are efficient if the worth of any
coalition can be calculated in polynomial time.
Here we use a sampling algorithm to estimate Shapley value which reduces 
the complexity to $\gO(mn)$, where $m$ is the number of samples taken
\cite{castro2009polynomial}.

Following the definition of Shapley value in \cref{eq:shapley}, 
an alternative definition of the Shapley value can be expressed in
terms of all possible orders of the players. Let $O: \{1,\dots,n\}
\rightarrow \{1,\dots,n\}$ be a permutation that assigns to each position 
$k$ the player $O(k)$. Let us denote by $\pi(\sP)$ the set of all possible 
permutations with player set $\sP$. Given a permutation $O$, we denote 
by $Pre^i(O)$ the set of predecessors of the player $i$ in the order $O$,
i.e. $Pre^i(O) = \{O(1),\dots,O(k-1)\}$, if $i=O(k)$.

It can be proved that the Shapley value in \cref{eq:shapley} can be 
expressed equivalently in the following way:
\begin{equation}
\label{eq:shapley-pre}
\begin{split}
Sh_i(f) = \frac{1}{n!} \sum_{O \in \pi(\sP)} 
\big[ f(Pre^i(O) \cup \{i\}) - f(Pre^i(O)) \big], \quad i = 1, \dots , n
\end{split}
\end{equation}
In estimation, we randomly take $m$ samples of player order $O$ from
$\pi(\sP)$, calculate the marginal contribution of the players in the order
$O$, which is defined in the summation of the equation above, and
finally average the marginal contributions as the approximation.

Then we will obtain, in polynomial time, an estimation
of the Shapley value with some desirable properties. To estimate
the Shapley value, we will use a unique sampling process for all
players. The sampling process is defined as follows:
\begin{itemize}
\item The population of the sampling process $P$ will be the set of all
possible orders of $n$ players. The vector parameter under study 
is $Sh = (Sh_1,\dots,Sh_n)$.
\item The characteristics observed in each sampling unit are 
the marginal contributions of the players in the order $O$, i.e.
\begin{equation}
\begin{split}
\chi(O) = \{\chi(O_i)\}_{i=1}^n, \quad where
\ \chi(O)_i = f(Pre^i(O) \cup \{i\}) - f(Pre^i(O)) 
\end{split}
\end{equation}
\item The estimate of the parameter will be the mean of the
marginal contributions over the sample $M$, i.e.
\begin{equation}
\begin{split}
\hat{Sh} = (\hat{Sh_1}, \dots, \hat{Sh_n}), \qquad where 
\quad \hat{Sh_i} = \frac{1}{m} \sum_{O \in M} \chi(O)_i.
\end{split}
\end{equation}
\end{itemize}

\begin{figure}[t]
\centering
\includegraphics[width=\linewidth]{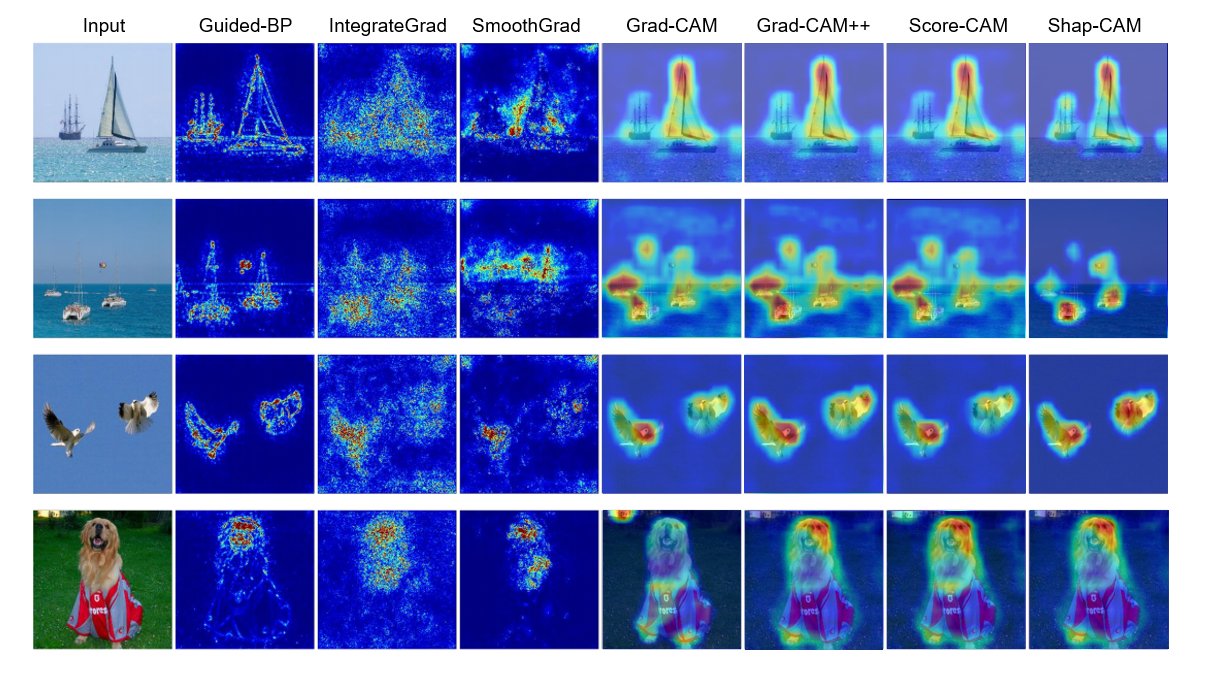}
\caption{Visualization results of 
Guided Backpropagation \cite{springenberg2014striving}, 
SmoothGrad \cite{smilkov2017smoothgrad}, 
IntegrateGrad \cite{sundararajan2017axiomatic},
Grad-CAM \cite{selvaraju2016gradcam},
Grad-CAM++ \cite{chattopadhay2018gradcam++}, 
Score-CAM \cite{wang2020scorecam}
and our proposed Shap-CAM.}
\label{fig:saliency}
\end{figure}

\section{Experiments}
\label{sec:experiment}

In this section, we conduct experiments to evaluate the
effectiveness of the proposed explanation method. We first introduce
the datasets for evaluation and our implementation details.
Then we assess the fairness of the explanation 
(the significance of the highlighted region for the model's decision)
qualitatively via visualization in \cref{sec:visualization}
and quantatively on image recognition in \cref{sec:recognition}. 
In \cref{sec:localization} we show the effectiveness for 
class-conditional localization of objects in a given image. 
The knowledge distillation experiment is followed in \cref{sec:kd}.

\subsection{Datasets and Implementation Details}

We first detail the datasets that we carried out experiments on:
The ImageNet (ILSVRC2012) dataset consists of about 1.2 million and 50k 
images from 1,000 classes for training and validation respectively.
We conducted the following experiments on the validation split of ImageNet.
The PASCAL VOC dataset consists of 9,963 natural images from 20 different 
classes. We used the PASCAL VOC 2007 trainval sets which contained 5,011 
images for the recognition evaluation.

For both the ImageNet and the PASCAL VOC datasets, all images are
resized to $224 \times 224 \times 3$, transformed to the range [0, 1], 
and then normalized using mean vector [0.485, 0.456, 0.406] and standard 
deviation vector [0.229, 0.224, 0.225]. In the following experiments, 
we use pre-trained VGG16 network \cite{simonyan2014very} from the Pytorch
model zoo as a base model. 
As for the calculation of Shapley value in \cref{eq:shap-cam}, 
only the pixels in the set $\sS$ are preserved and the others are set to 
the average value of the whole feature map.
Unless stated otherwise, the sampling number for 
estimating Shapley value is set to $10^4$. 
For a fair comparison, all saliency
maps are upsampled with bilinear interpolate to 224 $\times$ 224.

\subsection{Qualitative Evaluation via Visualization}
\label{sec:visualization}

We qualitatively compare the saliency maps produced by recently 
SOTA methods, including gradient-based methods 
(Guided Backpropagation \cite{springenberg2014striving}, 
IntegrateGrad \cite{sundararajan2017axiomatic}, 
SmoothGrad \cite{smilkov2017smoothgrad}), 
and activation-based methods (Grad-CAM \cite{selvaraju2016gradcam}, 
Grad-CAM \cite{selvaraju2016gradcam},
Grad-CAM++ \cite{chattopadhay2018gradcam++}) 
to validate the effectiveness of Shap-CAM.
As shown in \cref{fig:saliency}, results in Shap-CAM, random
noises are much less than that other methods. 
In addition, Shap-CAM generates 
smoother saliency maps comparing with gradient-based methods.
Due to the introduction of Shapley value, our Shap-CAM is able to estimate
the marginal contribution of the pixels and take the relationship between 
pixels into consideration, and thus can obtain more rational and accurate 
saliency maps.

\subsection{Faithfulness Evaluation via Image Recognition}
\label{sec:recognition}

The faithfulness evaluations are carried out as depicted
in Grad-CAM++ \cite{chattopadhay2018gradcam++} for the purpose of 
object recognition.
The original input is masked by point-wise multiplication with the 
saliency maps to observe the
score change on the target class. In this experiment, rather
than do point-wise multiplication with the original generated saliency 
map, we slightly modify by limiting the number of positive pixels 
in the saliency map.
Two metrics called Average Drop, Average Increase In
Confidence are introduced:

\textbf{Average Drop:} The Average Drop refers to the
maximum positive difference in the predictions made by the
prediction using the input image and the prediction using the
saliency map. It is given as: $\sum_{i=1}^N 
\dfrac{\max (0, Y_i^c-O_i^c)}{Y_i^c} \times 100\%$.
Here, $Y_i^c$ refers to the prediction score on class $c$ using the
input image $i$ and $O_i^c$ refers to the prediction score on class $c$
using the saliency map produced over the input image $i$.
A good explanation map for a class
should highlight the regions that are most relevant for
decision-making. It is expected that removing parts of an
image will reduce the confidence of the model in its decision, 
as compared to its confidence when the full image is provided as input.

\textbf{Increase in Confidence:} 
Complementary to the
previous metric, it would be expected that there must be
scenarios where providing only the explanation map region
as input (instead of the full image) rather increases the
confidence in the prediction (especially when the context
is distracting).
In this metric, we measure the number of
times in the entire dataset, the model's confidence increased
when providing only the explanation map regions as input. 
The Average Increase in
Confidence is denoted as:$\sum_{i=1}^N 
\dfrac{sign(Y_i^c < O_i^c)}{N} \times 100\%$, where
$sign$ presents an indicator function that returns 1 if input is True.

\begin{table}[t]
\caption{Recognition evaluation results on the ImageNet (ILSVRC2012)
validation set (lower is better in Average 
Drop, higher is better in Average Increase).}
\label{tab:recognition-table1}
\centering
\resizebox{\textwidth}{7mm}{
\begin{tabular}{|l|cccccc|}
\hline
Method  & \makebox[0.1\textwidth][c]{Mask} & \makebox[0.1\textwidth][c]{RISE} & 
\makebox[0.15\textwidth][c]{GradCAM} & \makebox[0.15\textwidth][c]{GradCAM++} & 
\makebox[0.15\textwidth][c]{ScoreCAM} & \makebox[0.12\textwidth][c]{ShapCAM} \\ 
\hline 
Avr. Drop(\%)      & 63.5 & 47.0 & 47.8 & 45.5 & 31.5 & \bf{28.0} \\
Avr. Increase(\%)  & 5.29 & 14.0 & 19.6 & 18.9 & 30.6 & \bf{31.8} \\
\hline
\end{tabular}}
\end{table}

\begin{table}[t]
\caption{Recognition evaluation results on the PASCAL VOC 2007
validation set (lower is better in Average 
Drop, higher is better in Average Increase).}
\label{tab:recognition-table2}
\centering
\resizebox{\textwidth}{7mm}{
\begin{tabular}{|l|cccccc|}
\hline
Method  & \makebox[0.1\textwidth][c]{Mask} & \makebox[0.1\textwidth][c]{RISE} & 
\makebox[0.15\textwidth][c]{GradCAM} & \makebox[0.15\textwidth][c]{GradCAM++} & 
\makebox[0.15\textwidth][c]{ScoreCAM} & \makebox[0.12\textwidth][c]{ShapCAM} \\ 
\hline
Avr. Drop(\%)      & 45.3 & 31.3 & 28.5 & 19.5 & 15.6 & \bf{13.2} \\
Avr. Increase(\%)  & 10.7 & 18.2 & 21.4 & 19.0 & 28.9 & \bf{32.7} \\
\hline
\end{tabular}}
\end{table}

Our comparison extends with state-of-the-art methods, namely gradient-based,
perturbation-based and CAM-based methods,
including Mask \cite{fong2017interpretable}, RISE \cite{petsiuk2018rise}, 
Grad-CAM \cite{selvaraju2016gradcam} and Grad-CAM++ 
\cite{chattopadhay2018gradcam++}.
Experiment conducts on the ImageNet (ILSVRC2012) validation
set, 2000 images are randomly selected. Results are reported in 
\cref{tab:recognition-table1}.
Results on the PASCAL VOC 2007 validation set are reported in 
\cref{tab:recognition-table2}.

As shown in \cref{tab:recognition-table1} and 
\cref{tab:recognition-table2}, Shap-CAM outperforms other perturbation-based 
and CAM-based methods. 
Shap-CAM can successfully locate the most distinguishable part of the 
target item, rather than only determining what humans think is important, 
based on its performance on the recognition challenge.
Results on the recognition task show that Shap-CAM can more accurately 
reveal the decision-making process of the original CNN model than 
earlier techniques. Previous methods lack solid theoretical 
foundation and suffer from self-designed explanations, which are prone to 
the manipulation of gradients or fail to
consider the relationship between different pixels. Our Shap-CAM instead
is able to obtain more rational and accurate contribution of each pixel
due to the superiority of Shapley value
and the consideration of relationship between pixels.

\begin{figure}[t]
\centering
\includegraphics[width=\linewidth]{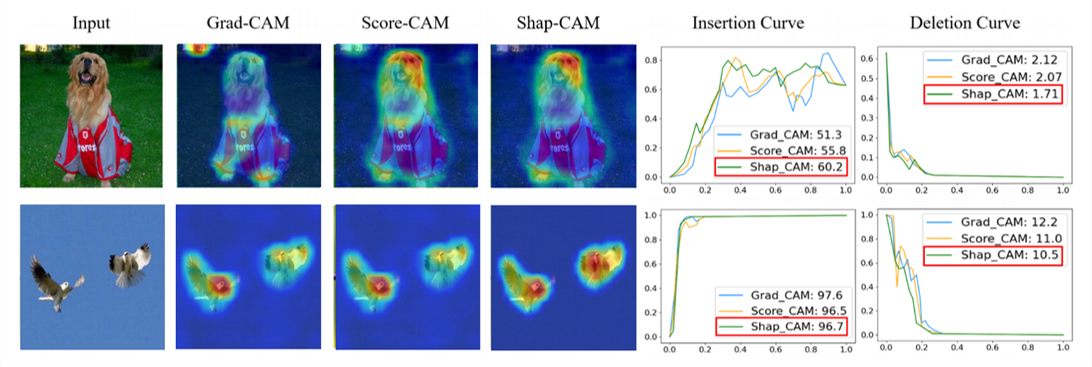}
\caption{Grad-CAM, Score-CAM and Shap-CAM generated saliency maps 
for representative images in terms of deletion and insertion
curves. In the insertion curve, a better explanation is expected 
that the prediction score to increase quickly, while in the deletion 
curve, it is expected the classification confidence to drop faster.}
\label{fig:insert_delete}
\end{figure}

Furthermore, for a more comprehensive comparison, we
also evaluate our method on deletion and insertion metrics which 
are proposed in \cite{petsiuk2018rise}.
In our experiment, we simply remove or introduce
pixels from an image by setting the pixel values to zero
or one with step 0.01 (remove or introduce 1\% pixels of
the whole image each step).
Example are shown in \cref{fig:insert_delete},
where our approach achieves better performance on both
metrics compared with SOTA methods.

\subsection{Localization Evaluation}
\label{sec:localization}

In this section, we measure the quality of the generated
saliency map through localization ability.
Bounding box evaluations are accomplished. 
We employ the similar metric, as specified in Score-CAM, called 
the Energy-based pointing game. Here, the
amount of energy of the saliency map is calculated by
finding out how much of the saliency map falls inside
the bounding box. Specifically, the input is binarized with
the interior of the bounding box marked as 1 and the region outside 
the bounding box as 0. Then, this input is
multiplied with the generated saliency map and summed
over to calculate the proportion ratio, which given as
\begin{align}
Proportion = \frac{\sum L_{(i,j) \in bbox}^c}
{\sum L_{(i,j) \in bbox}^c + \sum L_{(i,j) \notin bbox}^c}.
\end{align}
Two pre-trained models, namely VGG-16 \cite{simonyan2014very}, 
ResNet18 \cite{he2016deep}, are
used to conduct the energy-based pointing game on the 2000
randomly chosen images from the ILSVRC 2012 Validation set.

We randomly select images from the validation set by removing images where 
object occupies more than 50\% of the
whole image. For convenience, we only consider these images with only 
one bounding box for target class. 
We experiment on 500 random selected images from the ILSVRC
2012 validation set. Evaluation results are reported in 
\cref{tab:localization-table},
which show that our method outperforms previous works.
This also confirms that the Shap-CAM-generated saliency map has fewer noises.
As is shown in the previous research \cite{heo2019fooling}, the 
state-of-the-art gradient-based 
interpreters can be easily fooled by manipulating the gradients with no 
noticeable modifications to the original images, 
proving that the SOTA methods are not robust enough to the noise.
On the contrary, our Shap-CAM can alleviate the effect of this problem
by estimating the importance of each pixel more accurately.

\begin{table}[t]
\caption{Localization Evaluations of Proportion (\%) using Energy-based 
Pointing Game (Higher the better).}
\label{tab:localization-table}
\centering
\begin{tabular}{|l|cccc|}
\hline
\makebox[0.13\textwidth][l]{Method} & \makebox[0.17\textwidth][c]{Grad-CAM} & 
\makebox[0.17\textwidth][c]{Grad-CAM++} & 
\makebox[0.17\textwidth][c]{Score-CAM} & \makebox[0.15\textwidth][c]{Shap-CAM}\\ 
\hline
VGG-16      & 39.95 & 40.16 & 40.10 & \bf{40.45} \\
ResNet18    & 40.90 & 40.85 & 40.76 & \bf{41.28} \\
\hline
\end{tabular}
\end{table}

\subsection{Learning from Explanations: Knowledge Distillation}
\label{sec:kd}

Following the knowledge distillation experiment settings in Grad-CAM++
\cite{chattopadhay2018gradcam++}, we show that in a constrained 
teacher-student learning setting, knowledge transfer to a shallow
student (commonly called knowledge distillation) is possible
from the explanation of CNN decisions generated by CAM methods.
We use Wide ResNets \cite{zagoruyko2016wide} for both the student and 
teacher networks.
We train a WRN-40-2 teacher network (2.2 M parameters) on the
CIFAR-10 dataset. In order to train a student WRN-16-2 network 
(0.7 M parameters), we introduce a modified
loss $L_{stu}$, which is a weighted combination of the
standard cross entropy loss $L_{CE}$ and an interpretability
loss $L_{interp}$.
\begin{align}
\label{eq:kd}
L_{stu}(c, W_s, W_t, I) = L_{CE}(c, W_s(I)) + 
\alpha ||L_s^c(W_s(I)) - L_t^c(W_t(I))||_2^2
\end{align}
where the first term represents the cross entropy loss and the second 
term represents the interpretability loss $L_{interp}$.
In the above equations, $I$ indicates the input image and $c$ stands for 
the corresponding output class label. $L^c$ is the explanations given by 
a certain interpreter. $\alpha$ is a hyper parameter that controls
the importance given to the interpretability loss.
$W_s$ denotes the weights of the student network, and $W_t$ the weights of
the teacher network. This equation above forces the student network
not only minimize standard cross-entropy loss for classification, but
also learn from the most relevant parts of a given image
used for making a decision from the teacher network, which is the influence
of the interpretability loss $L_{interp}$ term.

\cref{tab:kd-table} shows the results for this experiment.
$L_{CE}$ is the normal cross
entropy loss function, i.e. the student network is trained
independently on the dataset without any intervention from
the expert teacher. The following four columns refer to loss functions
defined in \cref{eq:kd} where the explanations for image $I$ are 
generated using the corresponding interpreter.
We further also included $L_{KD}$, the knowledge distillation
loss introduced by Hinton et al. with temperature
parameter set to 4 \cite{hinton2015distilling}. 
It is indicated from these results that knowledge distillation can be 
improved by considering the explanations of the teacher.
The results also show that Shap-CAM provides better
explanation-based knowledge distillation than existing CAM-based
methods.

\begin{table}[t]
\caption{Test error rate (\%) for knowledge distillation to train a student
from a deeper teacher network. $L_{CE}$ is the normal cross
entropy loss function. The Column 2-6 refer to the modified loss function
$L_{stu}$ where the explanations for images are 
generated using the corresponding interpreter.}
\label{tab:kd-table}
\centering
\begin{tabular}{|l|ccccc|}
\hline
\makebox[0.17\textwidth][l]{Loss function} & \makebox[0.1\textwidth][c]{$L_{CE}$} & 
\makebox[0.15\textwidth][c]{GradCAM} & 
\makebox[0.15\textwidth][c]{GradCAM++} & 
\makebox[0.15\textwidth][c]{ScoreCAM} & \makebox[0.13\textwidth][c]{ShapCAM} \\ 
\hline
w/o $L_{KD}$ & 6.78 & 6.86 & 6.74 & 6.75 & \bf{6.69} \\
\hline
w/ $L_{KD}$ & 5.68 & 5.80 & 5.56 & 5.42 & \bf{5.37} \\
\hline
\end{tabular}
\end{table}

\section{Conclusion}

We propose Shap-CAM, a novel CAM variant, for visual explanations
of deep convolutional networks. 
We introduce Shapley value to represent the marginal contribution of 
each pixel to the model output.
Due to the superiority of Shapley value and the consideration of 
relationship between pixels, more rational and accurate explanations 
are obtained. We present evaluations of  the generated saliency
maps on recognition and localization tasks and show that 
Shap-CAM better discovers important features.
In a constrained teacher-student setting, 
our Shap-CAM provides better explanation-based knowledge distillation
than the state-of-the-art explanation approaches.

\section*{Acknowledgement}

This work was supported in part by the National Key Research and Development Program of China under Grant 2017YFA0700802, in part by the National Natural Science Foundation of China under Grant 62125603 and Grant U1813218, in part by a grant from the Beijing Academy of Artificial Intelligence (BAAI).

\clearpage
%
%
\bibliographystyle{splncs04}
\bibliography{egbib}
\end{document}